\documentclass{article}

\usepackage[preprint]{neurips_2025}

\usepackage[utf8]{inputenc}
\usepackage[T1]{fontenc}
\usepackage{hyperref}
\usepackage{url}
\usepackage{booktabs}
\usepackage{amsfonts}
\usepackage{nicefrac}
\usepackage{xcolor}
\usepackage{graphicx}
\usepackage{subcaption}
\usepackage{amsmath}
\usepackage{amssymb}
\usepackage{multirow}
\usepackage{array}
\usepackage{wrapfig}

\title{DevoTG: Temporal Graph Neural Networks for\\ Modeling C.\ elegans Developmental Connectomics}

\author{%
  Jayadratha Gayen$^{1,\ast}$ \qquad Bradly Alicea$^{2,3}$\\[6pt]
  $^{1}$IIIT Hyderabad \qquad
  $^{2}$OpenWorm Foundation \qquad
  $^{3}$Orthogonal Research and Education Laboratory\\[6pt]
  $^{\ast}$\texttt{contactjayag@gmail.com} \qquad \texttt{bradly.alicea@outlook.com}
}
\begin{document}

\maketitle

\begin{abstract}
Understanding how a nervous system wires itself from birth to adulthood is a fundamental challenge in developmental neuroscience.
We present \textbf{DevoTG}, a temporal graph framework that applies Temporal Graph Neural Networks (TGNs) to two complementary representations of \textit{C.~elegans} neural development: a Continuous-Time Dynamic Graph (CTDG) of cell division events derived from cell lineage data, and a Discrete-Time Dynamic Graph (DTDG) of the developing synaptic connectome spanning eight reconstructed electron-microscopy datasets.
On the lineage prediction task, our TGN achieves a mean test AUC of
$\mathbf{0.839 \pm 0.007}$ (5 seeds; validation AUC $0.937 \pm 0.001$),
outperforming a static GNN with the identical architecture by \textbf{26 AUC points} ($0.577 \pm 0.080$), demonstrating that temporal memory is the decisive factor.
Applied to the connectome DTDG, DevoTG identifies \textbf{three connection stability classes} (stable, developmental, and variable) across 225 neurons and 858 to 2{,}496 connections over development (L1 birth to adult), providing a temporal-graph-theoretic complement to the individual-variability classification of Witvliet et al.
Analysis of hub command interneurons AVA, AVB, and AVE reveals their
persistent centrality and how their integration roles are progressively
reinforced across larval stages. Accompanying interactive visualizations (3D animated networks, centrality heatmaps, and a spatiotemporal lineage graph) make developmental dynamics accessible for biological hypothesis generation.
DevoTG is open-source and designed for extension to other developing
nervous systems. Code is publicly available at \url{https://github.com/DevoLearn/DevoGraph/tree/main/DevoTG}.
\end{abstract}

\section{Introduction}
\label{sec:intro}
Neural circuit formation is a dynamic, self-organizing process in which
billions of molecular signals, cell divisions, and axonal migrations collectively
produce a functional nervous system.
Understanding the rules governing which connections form, which persist,
and which are pruned is a core question in developmental neuroscience.
The nematode \textit{Caenorhabditis elegans} offers an unmatched opportunity
to study these rules at cellular resolution: its invariant cell lineage is
entirely mapped, its nervous system contains a tractable number of neurons,
and the complete adult connectome has been characterized at synaptic
precision~\cite{white1986structure}.

A landmark study by Witvliet et al.~\cite{witvliet2021connectomes}
extended this foundation by reconstructing the full brains of eight isogenic
\textit{C.~elegans} individuals across postnatal developmental stages using
serial-section electron microscopy.
Their work revealed that while overall brain geometry is preserved from birth,
substantial rewiring occurs: approximately 4{,}500 new synapses strengthen
existing connections, and 1{,}200 new synaptic connections form by adulthood.
Crucially, around 43\% of all cell-cell connections are not conserved between
individuals, challenging the classical view of the \textit{C.~elegans}
connectome as hardwired~\cite{witvliet2021connectomes}.

Despite these advances, the computational modeling of \textit{C.~elegans}
neural development has largely treated each developmental snapshot as a
separate static graph~\cite{towlson2013caenorhabditis, bassett2017network}.
This framing discards the temporal ordering and continuous dynamics that
drive developmental change.
Temporal Graph Neural Networks (TGNs)~\cite{rossi2020temporal, trivedi2019dyrep},
which maintain evolving node memories over a stream of timestamped events,
offer a natural representational fit for biological development: each cell
division or synapse formation event updates the system state, and future
connectivity depends on this accumulated history.

We present \textbf{DevoTG} (\textbf{Devo}lopmental \textbf{T}emporal \textbf{G}raph),
a framework that applies temporal graph methods to two complementary
representations of \textit{C.~elegans} development.
The first component models cell lineage as a Continuous-Time Dynamic Graph
(CTDG): cell division events are timestamped edges, and a TGN learns to predict
which parent cell divides into which daughter cells, capturing developmental
timing and spatial context.
The second component constructs a Discrete-Time Dynamic Graph (DTDG) of the
synaptic connectome from the eight Witvliet datasets, enabling systematic
analysis of how connection topology evolves from birth to adulthood.
Together, these components integrate two fundamental axes of \textit{C.~elegans}
development (lineage and connectivity) under a unified temporal graph framework.

\paragraph{Contributions.} This work makes four main contributions:
\begin{enumerate}
  \item \textbf{DevoTG framework}: An integrated pipeline for temporal graph
        analysis of \textit{C.~elegans} development, combining CTDG-based
        lineage modeling and DTDG-based connectome analysis.
  \item \textbf{TGN for cell division prediction}: The first application of
        Temporal Graph Neural Networks to C. elegans cell lineage prediction,
        achieving a test AUC of $\mathbf{0.839 \pm 0.007}$ (mean $\pm$ std,
        5 seeds), 26 points above a static GNN baseline with identical
        architecture ($0.577 \pm 0.080$), demonstrating that temporal memory
        is the decisive inductive bias for developmental link prediction.
  \item \textbf{Temporal connection stability analysis}: A three-class taxonomy
        of synaptic connections (stable, developmental, variable) derived from
        temporal connectivity patterns across 8 developmental timepoints,
        providing a time-series-based complement to Witvliet's
        individual-variability classification.
  \item \textbf{Interactive visualizations}: A suite of tools (3D animated
        network dynamics, centrality heatmaps, circuit-level views of
        command interneurons, and a spatiotemporal lineage graph) designed
        to enable biological hypothesis generation from temporal connectome data.
\end{enumerate}

\section{Related Work}
\label{sec:related}
\subsection*{C. elegans Connectomics and Developmental Neuroscience}

The foundational map of the adult \textit{C.~elegans} hermaphrodite nervous
system by White et al.~\cite{white1986structure} established 302 neurons and
~7{,}000 chemical and electrical synapses, providing the first complete wiring
diagram of any nervous system.
Subsequent studies refined this map using modern electron microscopy
techniques: Varshney et al.~\cite{varshney2011structural} produced a
large-scale reconstruction of the hermaphrodite somatic nervous system,
and Cook et al.~\cite{cook2019whole} extended coverage to both sexes.
The Witvliet et al.~\cite{witvliet2021connectomes} study introduced the
critical developmental dimension, reconstructing eight isogenic individuals
from birth to adulthood and establishing quantitative principles of synaptic
maturation that DevoTG is designed to analyze computationally.

From a developmental perspective, Alicea and Gordon~\cite{alicea2018spatialnetworks}
framed cell differentiation as spatial networks, and Alicea~\cite{alicea2020raising}
characterized the emergence of neuronal activity and behavior in
\textit{C.~elegans} through network principles.
Frankel and Kurshan~\cite{frankel2025devsynapse} reviewed molecular mechanisms
of synaptogenesis in \textit{C.~elegans}, providing biological context for the
connectivity patterns that DevoTG quantifies.
Our work operationalizes this developmental biology through temporal graph
representations, enabling data-driven discovery at scale.

\subsection*{Temporal Graph Neural Networks}

Static GNNs (e.g., GCN, GAT) aggregate neighborhood information over fixed
graphs but cannot represent the temporal order of events~\cite{longa2023graph}.
Several architectures have been proposed for dynamic graphs.
DyRep~\cite{trivedi2019dyrep} uses a latent representation model that updates
node embeddings upon each interaction event, applied to social and collaboration
networks.
JODIE~\cite{kumar2020predicting} learns coupled user-item trajectory embeddings
for temporal interaction prediction.
The Temporal Graph Network (TGN)~\cite{rossi2020temporal} unifies these
approaches with a modular design: a memory module that stores a compressed
summary of each node's interaction history, a message function that processes
event features, an attention-based aggregation mechanism, and a graph embedding
module.
TGN has demonstrated state-of-the-art performance on temporal link prediction
benchmarks including Reddit, Wikipedia, and Twitter social networks.

Variational Graph Recurrent Neural Networks (VGRNN)~\cite{hajiramezanali2019variational}
model latent structural dynamics and are particularly suited to sparse,
noisy temporal graphs.
EvolveGCN~\cite{pareja2020evolvegcn} evolves GCN weight matrices over time using
RNNs, operating on discrete graph snapshots.
DevoTG's CTDG component employs TGN because it is the most natural fit for
cell division data: each division event is a timestamped interaction, and
the accumulated developmental history of each cell is precisely what the TGN
memory module is designed to capture.

\subsection*{Dynamic Network Analysis in Neuroscience}

The field of network neuroscience~\cite{bassett2017network} has applied graph
theoretic methods to understand brain organization, including degree distributions,
small-world properties~\cite{watts1998collective}, community structure, and hub
neuron classification~\cite{towlson2013caenorhabditis}.
Holme and Saramäki~\cite{holme2012temporal} established a framework for
temporal networks in biology, distinguishing between node-level temporal
properties and edge-level activity patterns.
Hosseinzadeh et al.~\cite{hosseinzadeh2022temporal} reviewed temporal network
models in biology and medicine, noting the gap between static connectome
analyses and the inherently dynamic processes that shape them.
DevoTG bridges this gap for the \textit{C.~elegans} connectome by applying
DTDG representations to the Witvliet longitudinal dataset.

Hub neuron analysis in \textit{C.~elegans} has identified AVA, AVB, AVE, and
related command interneurons as disproportionately connected nodes that
integrate sensory input and drive locomotion decisions~\cite{towlson2013caenorhabditis}.
Our Section~\ref{subsec:circuit} provides a temporal graph perspective on how
this hub status is maintained and reinforced across development.

\subsection*{Visualization of Temporal Networks}

Beck et al.~\cite{beck2017visual} provide a comprehensive taxonomy of dynamic
graph visualization techniques, categorizing approaches along animation,
timeline, and 3D space dimensions.
Rosvall and Bergstrom~\cite{rosvall2010mapping} proposed alluvial diagrams for
mapping change in large community structures over time.
DevoTG's interactive visualization suite draws on these design principles to
build Plotly-based 3D animations of the developing connectome, node importance
heatmaps over developmental time, and circuit-level ego-network views that are
navigable by neuroscientists without programming expertise.

\section{Methods}
\label{sec:methods}
\subsection{Data Sources}
\label{subsec:data}

DevoTG integrates two datasets that capture complementary aspects of
\textit{C.~elegans} neural development.

\paragraph{Cell lineage dataset.}
We use a curated CSV of cell division events derived from WormAtlas canonical
lineage data.
Each row records a parent cell, its 3D spatial coordinates
$(x, y, z)$ in microns, the names of two daughter cells, and the birth time
in minutes post-fertilization.
The dataset contains 642 division events spanning 1{,}203 unique cell states
from the zygote (P0) through approximately five generations of division,
covering the period from fertilization to early larval development.
Birth times range from 0 to $\sim$600 minutes, capturing the embryonic and
early post-embryonic periods.

\paragraph{Developmental connectome dataset.}
We use the eight-timepoint connectome dataset from Witvliet et al.~\cite{witvliet2021connectomes},
downloaded from the ConnectomeToolbox repository.
Each dataset corresponds to an isogenic \textit{C.~elegans} hermaphrodite at
a distinct developmental stage: four L1 larvae (0 h, 5 h, 8 h, and 16 h
post-hatching), one L2 larva (23 h), one L3 larva (27 h), and two young adults
(both at 45 h).
The processed DTDG covers 225 neurons across all timepoints (sensory: 64,
interneuron: 43, motor: 42, muscle: 32, modulatory: 31, glia: 10, unknown: 3)
and includes both chemical synapses and gap junctions.

\subsection{Temporal Graph Representations}
\label{subsec:representations}

\paragraph{CTDG for cell lineage.}
We represent each cell division event as a directed, timestamped edge in a
Continuous-Time Dynamic Graph $\mathcal{G} = (V, E, \mathcal{T})$, where
$V$ is the set of cell states, $E$ is the set of division events, and
$\mathcal{T} \subseteq \mathbb{R}_{+}$ is event time.
Each division produces two edges: parent $\rightarrow$ daughter\,1 and
parent $\rightarrow$ daughter\,2, both at the daughter's birth time.
Node feature vectors $\mathbf{x}_i \in \mathbb{R}^{172}$ encode 3D position
(3 dims), birth time (1 dim), generation depth (1 dim), and zero-padding;
edge feature vectors $\mathbf{m}_{ij} \in \mathbb{R}^{172}$ encode the
spatial displacement between parent and daughter cell centroids.
This results in 1{,}203 nodes and 1{,}284 events in the CTDG.

\paragraph{DTDG for connectome.}
The synaptic connectome is represented as a Discrete-Time Dynamic Graph
$\mathcal{G} = (V, \{G_t\}_{t=1}^{8})$, where each $G_t = (V, E_t, W_t)$
is a snapshot of the connectome at timepoint $t$.
Edge weights $W_t$ are synapse counts.
We retain connections with $\geq 1$ synapse (chemical or gap junction), yielding
edge counts that grow from 858 at birth to 2{,}496 in the adult (Table~\ref{tab:growth}).

\begin{table}[h]
\centering
\caption{Connectome growth across development (DevoTG DTDG).}
\label{tab:growth}
\small
\begin{tabular}{lrrrrrr}
\toprule
Stage & Time (h) & Edges & Chemical & Electrical & Synapses & Density \\
\midrule
L1 (birth)  & 0  & 858   & 775   & 83  & 1{,}400 & 0.017 \\
L1          & 5  & 1{,}110 & 986 & 124 & 2{,}028 & 0.022 \\
L1          & 8  & 1{,}106 & 1{,}012 & 94 & 2{,}237 & 0.022 \\
L1          & 16 & 1{,}345 & 1{,}136 & 209 & 3{,}020 & 0.027 \\
L2          & 23 & 1{,}807 & 1{,}515 & 292 & 4{,}535 & 0.036 \\
L3          & 27 & 1{,}739 & 1{,}525 & 214 & 4{,}723 & 0.035 \\
Adult       & 45 & 2{,}493 & 2{,}202 & 291 & 7{,}867 & 0.049 \\
Adult       & 45 & 2{,}496 & 2{,}186 & 310 & 8{,}400 & 0.050 \\
\bottomrule
\end{tabular}
\end{table}

\subsection{Temporal Graph Neural Network Architecture}
\label{subsec:tgn}

Our TGN implementation follows Rossi et al.~\cite{rossi2020temporal} and is
built on PyTorch Geometric.
It comprises three modules:

\paragraph{Memory module.}
A \texttt{TGNMemory} module maintains a memory vector
$\mathbf{m}_i \in \mathbb{R}^{100}$ per node, updated via a GRU cell when
node $i$ participates in a division event:
\begin{equation}
  \mathbf{m}_i(t^+) = \mathrm{GRU}\!\left(\mathbf{m}_i(t^-),\,
  [\mathbf{m}_j \oplus \mathbf{e}_{ij} \oplus \phi(t)]\right),
\end{equation}
where $\mathbf{e}_{ij}$ is the edge feature, $\phi(t)$ is a learned time
encoding with $d_t = 100$ dimensions, and $\oplus$ denotes concatenation.
The message function is identity (IdentityMessage) and the aggregator retains
the most recent message (LastAggregator).
The memory module has 103{,}200 trainable parameters.

\paragraph{Graph attention embedding.}
For a node $i$ and its sampled temporal neighborhood $\mathcal{N}_i$,
embeddings are computed by:
\begin{equation}
  \mathbf{z}_i = \mathrm{TransformerConv}\!\left(\mathbf{m}_i,\;
  \{(\mathbf{m}_j, [\phi(\Delta t_{ij}) \oplus \mathbf{e}_{ij}])\}_{j \in \mathcal{N}_i}\right),
\end{equation}
using a 2-head \texttt{TransformerConv} layer (input: 100, output: 100,
dropout 0.1).
Relative times $\Delta t_{ij} = t_{\text{last},i} - t_{ij}$ are encoded via
the shared time encoder.

\paragraph{Link predictor.}
For a candidate edge $(i, j)$, the prediction score is:
\begin{equation}
  \hat{y}_{ij} = \sigma\!\left(\mathbf{W}_3 \,\mathrm{ReLU}\!\left(
  \mathbf{W}_1\mathbf{z}_i + \mathbf{W}_2\mathbf{z}_j\right)\right),
\end{equation}
where $\mathbf{W}_1, \mathbf{W}_2 \in \mathbb{R}^{100 \times 100}$ and
$\mathbf{W}_3 \in \mathbb{R}^{1 \times 100}$.

\paragraph{Training.}
We train with Adam (lr = 0.001) minimizing binary cross-entropy on positive
edges and randomly sampled negative edges (ratio 1:1).
The data is split temporally: 70\% of events for training, 15\% for validation,
15\% for testing, ensuring the model is evaluated on strictly future events.
We train for 20 epochs with batch size 200, using 10 temporal neighbors per
node (LastNeighborLoader).
Total trainable parameters: 132{,}501.

\subsection{DTDG Analysis Pipeline}
\label{subsec:dtdg}

\paragraph{Connection stability classification.}
We classify each unique (source, target, type) triplet across the 8 timepoints
into one of three stability classes based on its temporal persistence:
\begin{itemize}
  \item \textbf{Stable}: present in $\geq 6$ of 8 timepoints (650 connections,
        15.1\%); represents core, persistent circuit architecture.
  \item \textbf{Developmental}: present in 2–5 timepoints (1{,}207 connections,
        28.1\%); appears transiently during specific larval stages, suggestive
        of scaffolding or transitional circuit roles.
  \item \textbf{Variable}: present in exactly 1 timepoint (2{,}440 connections,
        56.8\%); highly transient, likely capturing measurement variability
        and weak connections.
\end{itemize}

\paragraph{Network topology analysis.}
For each snapshot $G_t$, we compute standard graph metrics using NetworkX:
degree and betweenness centrality for all nodes, graph density, average path
length (on the largest connected component), and community structure via the
Louvain algorithm.
Node importance is tracked across all eight timepoints to identify hub neurons
and characterize their developmental trajectories.

\section{Results}
\label{sec:results}
\subsection{A Sample DevoTG: Topology Across Developmental Time}
\label{subsec:sample}

\begin{figure}[t]
  \centering
  \includegraphics[width=\linewidth]{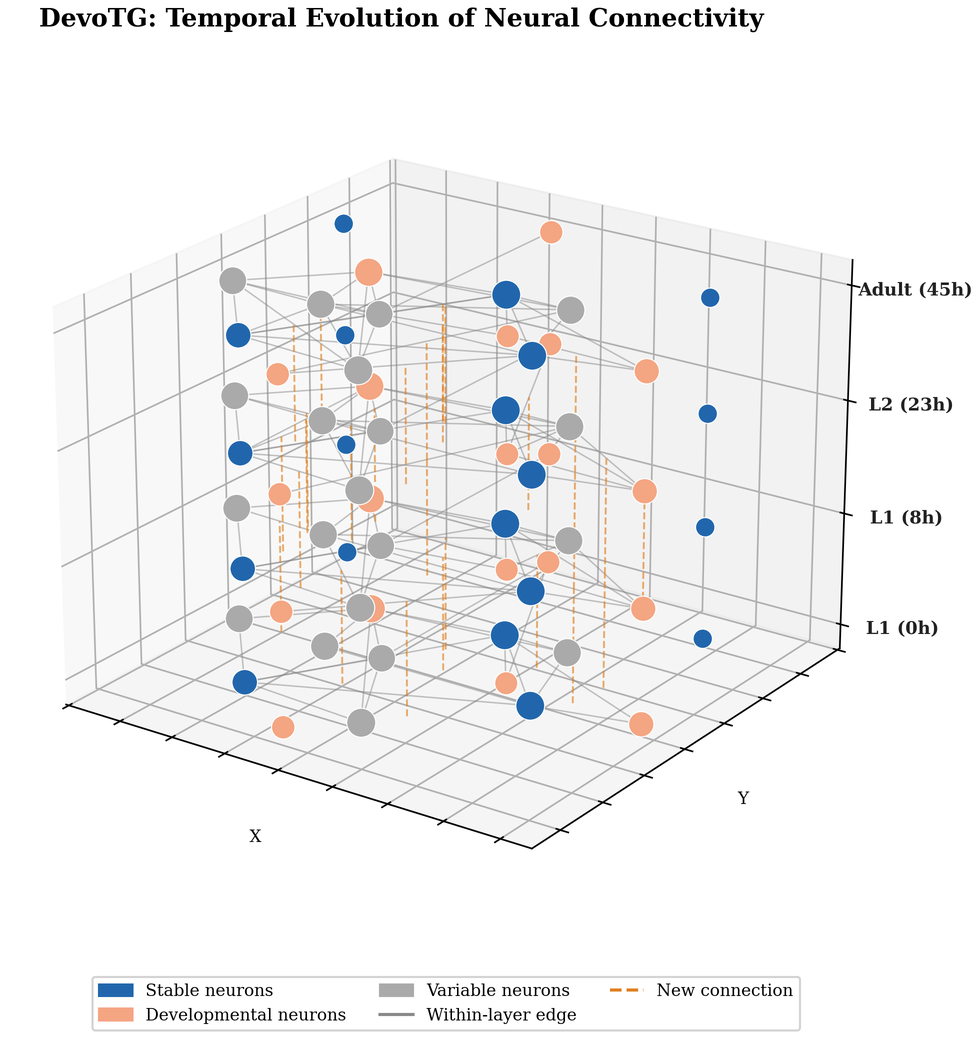}
  \caption{\textbf{Sample DevoTG: temporal evolution as staggered layers.}
    The 15 most-connected neurons are shown at four timepoints (L1 birth,
    L1 16 h, L2, adult) stacked along the $z$-axis.
    Orange dashed edges indicate connections that appear for the first time
    between adjacent timepoints.
    Node color: blue = stable class, salmon = developmental class,
    gray = variable class.}
  \label{fig:sample}
\end{figure}

Figure~\ref{fig:sample} illustrates the core representational idea of DevoTG
applied to the 15 most highly connected neurons in the connectome.
The network is shown at four developmental timepoints (L1 birth, L1 16 h, L2,
and Adult), stacked as layers along the $z$-axis so that topology changes
become spatially visible.
Within each layer, connections are drawn in gray; newly appearing connections
between adjacent layers are highlighted in orange.
Nodes are colored by their temporal stability class (blue: stable, salmon:
developmental, gray: variable).

Several patterns are immediately visible.
The core stable (blue) neurons maintain consistent positions and high degree
across all layers, forming a persistent topological backbone.
Developmental (salmon) nodes show peak connectivity in middle layers (L1 16 h
and L2) before either stabilizing or dropping out.
The density of inter-layer orange edges is highest in the L1$\rightarrow$L2
transition, consistent with the quantitative finding that edge count grows
most rapidly during this period (858$\rightarrow$1{,}807, a 110\% increase;
Table~\ref{tab:growth}).
This staggered-layer representation, inspired by hypergraph visualizations in
Alicea et al.~\cite{alicea2018spatialnetworks}, enables direct visual
inspection of the wiring changes that underlie maturation.

\subsection{Connection Stability: DevoTG vs.\ Witvliet et al.}
\label{subsec:stability}

\begin{figure}[t]
  \centering
  \includegraphics[width=\linewidth]{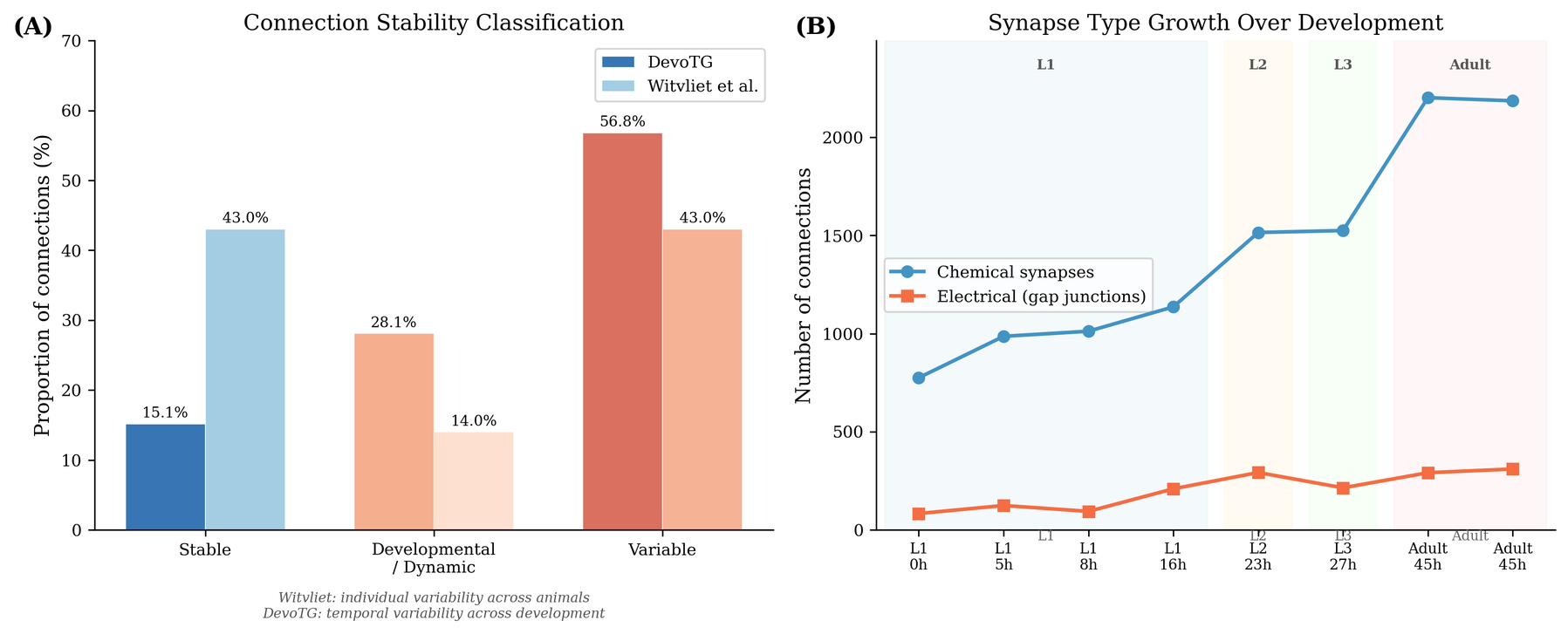}
  \caption{\textbf{Connection stability analysis.}
    (A) Proportion of connections per stability class for DevoTG (temporal
    classification, this work) and Witvliet et al.\ (individual variability
    classification~\cite{witvliet2021connectomes}).
    Note that the two frameworks measure different axes of variability.
    (B) Edge count over developmental time, split by synapse type.
    Chemical synapses grow monotonically; electrical connections show a
    less regular trajectory.}
  \label{fig:stability}
\end{figure}

Figure~\ref{fig:stability}A compares the proportion of connections in each
stability class between DevoTG's temporal classification and the
individual-variability classification reported by Witvliet et al.
The two frameworks operate on different axes: DevoTG classifies connections
by their temporal persistence across 8 developmental timepoints of a single
animal series, while Witvliet classify by whether connections are conserved
across isogenic individuals. Together they characterize how the
\textit{C.~elegans} connectome balances stereotypy and flexibility.

In DevoTG, 56.8\% of unique connection pairs appear in only a single
timepoint (variable class), 28.1\% appear transiently across 2–5 timepoints
(developmental class), and 15.1\% persist in $\geq$6 of 8 timepoints
(stable class).
In Witvliet et al., approximately 43\% of connections are variable
(individual-specific), while the remainder are either stable ($\sim$43\%) or
developmentally dynamic ($\sim$14\%) across individuals.
The higher proportion of temporally variable connections in DevoTG may reflect
two factors: (i) DevoTG includes gap junctions (excluded from Witvliet's
primary classification), which are more variable; and (ii) many
single-timepoint connections correspond to weak connections (weight=1) that
fluctuate around the detection threshold.

Figure~\ref{fig:stability}B shows the growth of chemical and electrical
connections separately over the eight timepoints.
Chemical synapses dominate throughout (775 to 2{,}202) and grow monotonically,
while electrical connections (gap junctions) show a more variable trajectory,
peaking at L1-16h (209) before plateauing.
This divergent growth pattern is biologically interpretable: chemical synapse
addition is the primary mechanism of connectome maturation, consistent with
Witvliet et al.'s observation that new chemical synapses account for the
majority of adult connectivity.

\subsection{Circuit-Level Analysis: AVA, AVB, and AVE Command Interneurons}
\label{subsec:circuit}

\begin{figure}[t]
  \centering
  \includegraphics[width=\linewidth]{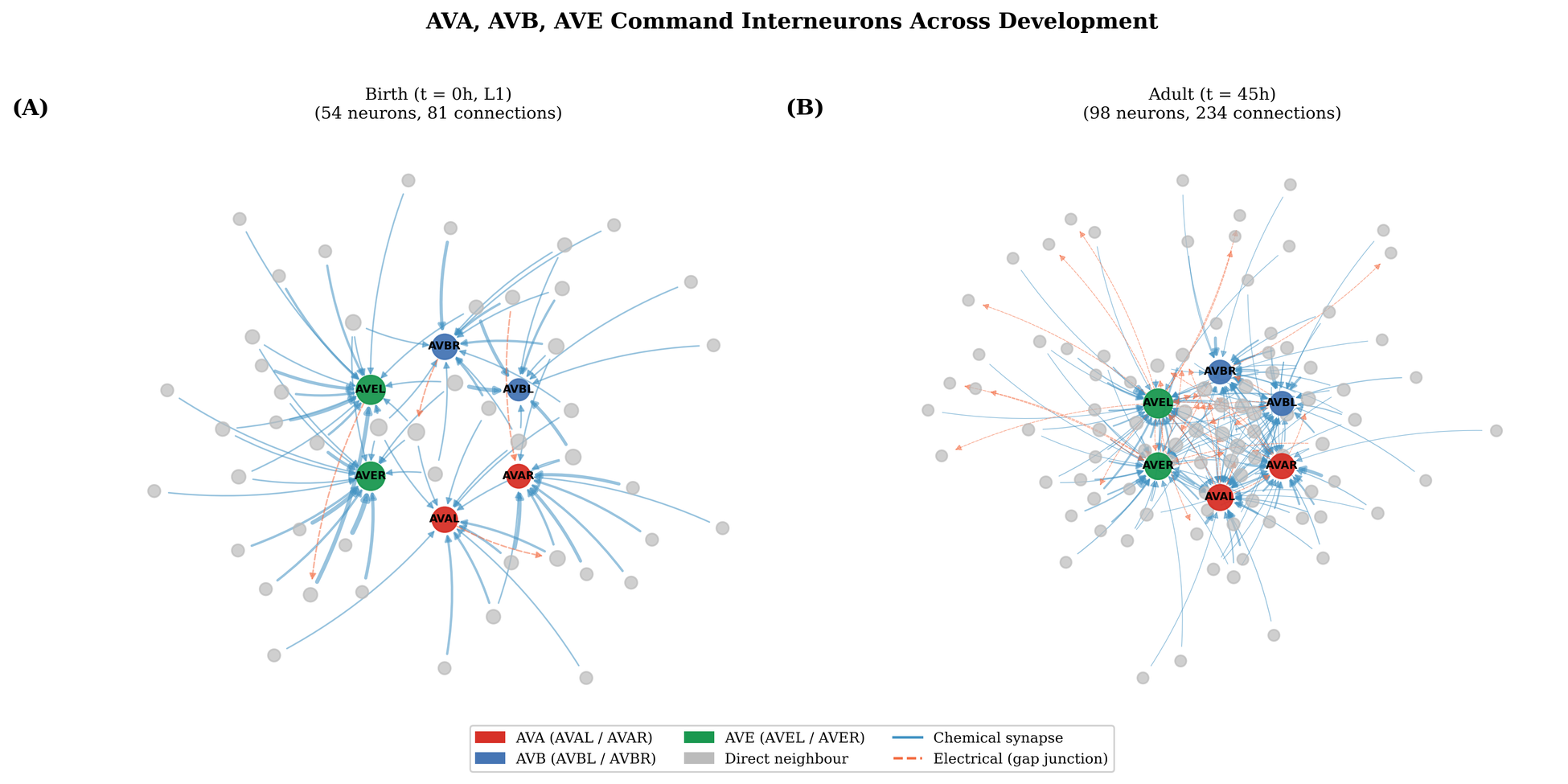}
  \caption{\textbf{AVA, AVB, AVE command interneurons in the developing connectome.}
    1-hop ego-networks at L1 birth (left) and adult (right).
    Node colors: red = AVA pair, blue = AVB pair, green = AVE pair,
    gray = connected neighbors.
    Solid edges: chemical synapses; dashed: gap junctions.
    Edge width $\propto$ synapse count.
    Node size $\propto$ degree.
    The core interneuron circuitry is present at birth; adulthood adds
    substantially more input connections from sensory and motor neurons.}
  \label{fig:circuit}
\end{figure}

The command interneurons AVA, AVB, and AVE (three bilateral pairs: AVAL/AVAR,
AVBL/AVBR, AVEL/AVER) are among the most studied neurons in
\textit{C.~elegans}~\cite{towlson2013caenorhabditis}.
They integrate sensory input and drive locomotion decisions (forward/backward),
placing them at the core of the animal's sensorimotor circuit.
Figure~\ref{fig:circuit} shows the 1-hop ego-networks of these six neurons at
birth (L1, 0 h) and in the adult (45 h), with chemical synapses as solid
edges and gap junctions as dashed edges, edge width proportional to synapse
weight, and node size proportional to degree.

Several features are notable.
First, the core connectivity among the six command interneurons themselves
is present from birth: AVAL and AVAR share both chemical and electrical
connections in L1, as do the AVB and AVE pairs.
This is consistent with Witvliet's finding that interneuron-interneuron
connections are predominantly stable; the decision-making architecture is
present at birth.
Second, by adulthood the number of direct neighbors (1-hop) of each command
interneuron increases substantially, with new chemical connections from motor
and sensory neurons adding to the integration capacity.
The betweenness centrality of AVAL, the highest-centrality node in the adult
graph (degree centrality 0.252), increases visibly as more neural pathways
route through it.
Third, AVE neurons add the most connections between L1 and adult, consistent
with their role in integrating mechanosensory inputs that mature postnatally.

\subsection{Spatiotemporal Development Graph}
\label{subsec:spatiotemporal}

\begin{figure}[t]
  \centering
  \includegraphics[width=\linewidth]{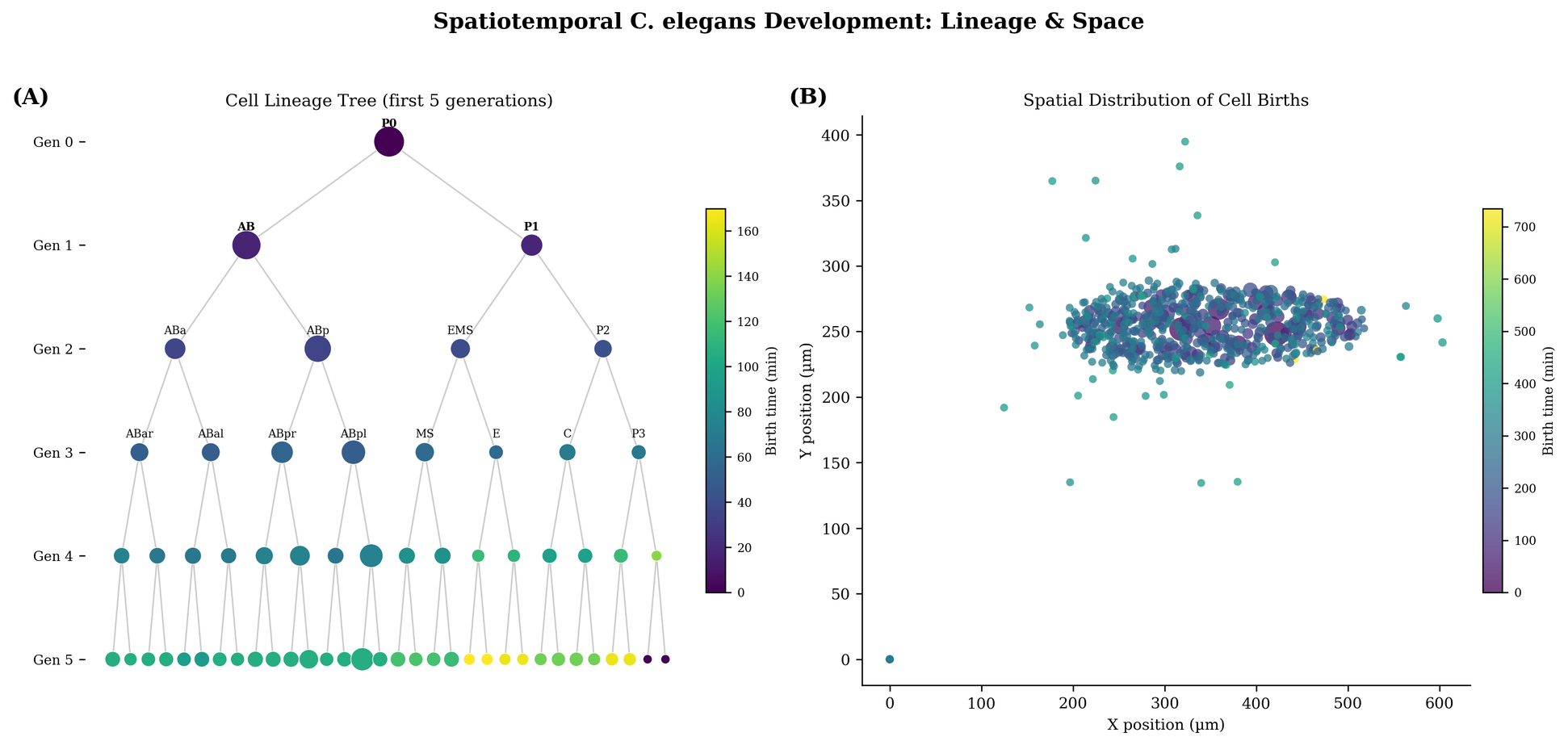}
  \caption{\textbf{Spatiotemporal development graph.}
    (A) Cell lineage tree for the first five generations, colored by birth
    time (viridis). Node size $\propto$ number of descendants.
    (B) Spatial scatter of all 642 parent cell positions, colored by birth
    time. An anterior-to-posterior gradient in $x$ and a core-to-periphery
    gradient in birth order are visible.}
  \label{fig:spatiotemporal}
\end{figure}

Figure~\ref{fig:spatiotemporal} presents a spatiotemporal development graph
that integrates cell lineage (the time axis of division) with the spatial
positions of cell births, analogous in spirit to the spatiotemporal hypergraph
visualizations in Alicea and Gordon~\cite{alicea2018spatialnetworks}.

Panel A shows the lineage tree for the first five generations of cell division,
rooted at the zygote P0 and growing outward.
Nodes are colored by birth time (viridis colormap, dark = early, bright = late)
and sized by number of descendants.
The canonical binary branching pattern of early \textit{C.~elegans} lineage
is clearly visible, with the AB lineage (anterior founder) dividing faster
than the P lineage (posterior stem).
Nodes at the tips of the tree correspond to cells that have ceased division
and are differentiating into specific cell types.

Panel B shows the spatial distribution of all 642 parent cell birth positions,
colored by birth time.
The anterior-posterior axis is visible as a gradient along $x$, and the
dorsal-ventral gradient along $y$.
Early cell births (dark, P0–AB generation) cluster near the center of the
embryo, while later births (bright) are distributed toward the periphery,
consistent with the inside-out pattern of \textit{C.~elegans} embryogenesis.

Together, Panels A and B allow researchers to trace any cell's lineage
identity (Panel A) and correlate it with its spatial position (Panel B),
enabling hypotheses about how lineage-position relationships constrain
eventual circuit membership, a direction highlighted in Alicea and
Gordon~\cite{alicea2018spatialnetworks} and extended by DevoTG's computational
framework.

\subsection{TGN Performance on Cell Division Prediction}
\label{subsec:performance}

Table~\ref{tab:baselines} reports the test-set performance of DevoTG's TGN
and three baseline models on the cell division prediction task, averaged over
\textbf{5 independent random seeds} (seeds 0–4).
For each run, test metrics are taken at the epoch with the highest validation
AUC (best-checkpoint evaluation), following standard ML practice.
All models use the same temporal split (70/15/15 train/val/test by event
order), with all sources of randomness (weight initialization and negative-edge
sampling) fully seeded for reproducibility.

\begin{table}[h]
  \centering
  \caption{Cell division link prediction performance (mean $\pm$ std, 5 seeds,
           evaluated at best-validation-AUC checkpoint).}
  \label{tab:baselines}
  \small
  \begin{tabular}{lcc}
    \toprule
    Model & Test AUC & Test AP \\
    \midrule
    Random baseline              & $0.539 \pm 0.000$ & $0.553 \pm 0.000$ \\
    Degree heuristic (PA)        & $0.463 \pm 0.012$ & $0.488 \pm 0.005$ \\
    Static GNN (GAT, no memory)  & $0.577 \pm 0.080$ & $0.563 \pm 0.061$ \\
    \midrule
    \textbf{DevoTG TGN (ours)}   & $\mathbf{0.839 \pm 0.007}$ & $\mathbf{0.763 \pm 0.006}$ \\
    \bottomrule
  \end{tabular}
\end{table}

The TGN achieves a test AUC of $\mathbf{0.839 \pm 0.007}$ and AP of
$\mathbf{0.763 \pm 0.006}$, substantially and consistently above all baselines,
with a standard deviation of 0.007 confirming stable, reproducible training.

The most informative comparison is against the Static GNN ($0.577 \pm 0.080$),
which uses the identical TransformerConv graph attention architecture but
replaces the TGN memory module with a standard static GNN over the accumulated
training graph.
The \textbf{26-point AUC gap} between Static GNN and TGN demonstrates that
\emph{temporal memory is the critical inductive bias} for developmental link
prediction: knowing not just which cells are currently connected, but
\emph{when} and in \emph{what order} connections formed, is essential for
predicting future division events.
The Static GNN's high variance ($\pm$0.080) additionally reveals that without
temporal context, models are sensitive to initialization, sometimes finding useful static features and
often not.

The degree heuristic (preferential attachment), scoring candidate edges by
$\sqrt{\deg(u) \cdot \deg(v)}$, achieves an AUC ($0.463$) \emph{below} random
($0.539$), which is informative: cell division is not preferential-attachment
driven.
Cells with high division counts in the training period are not systematically
the ones that divide most in the test period; \textit{C. elegans} lineage
follows a fixed developmental program, not an activity-driven attachment rule.

\begin{figure}[t]
  \centering
  \includegraphics[width=0.48\linewidth]{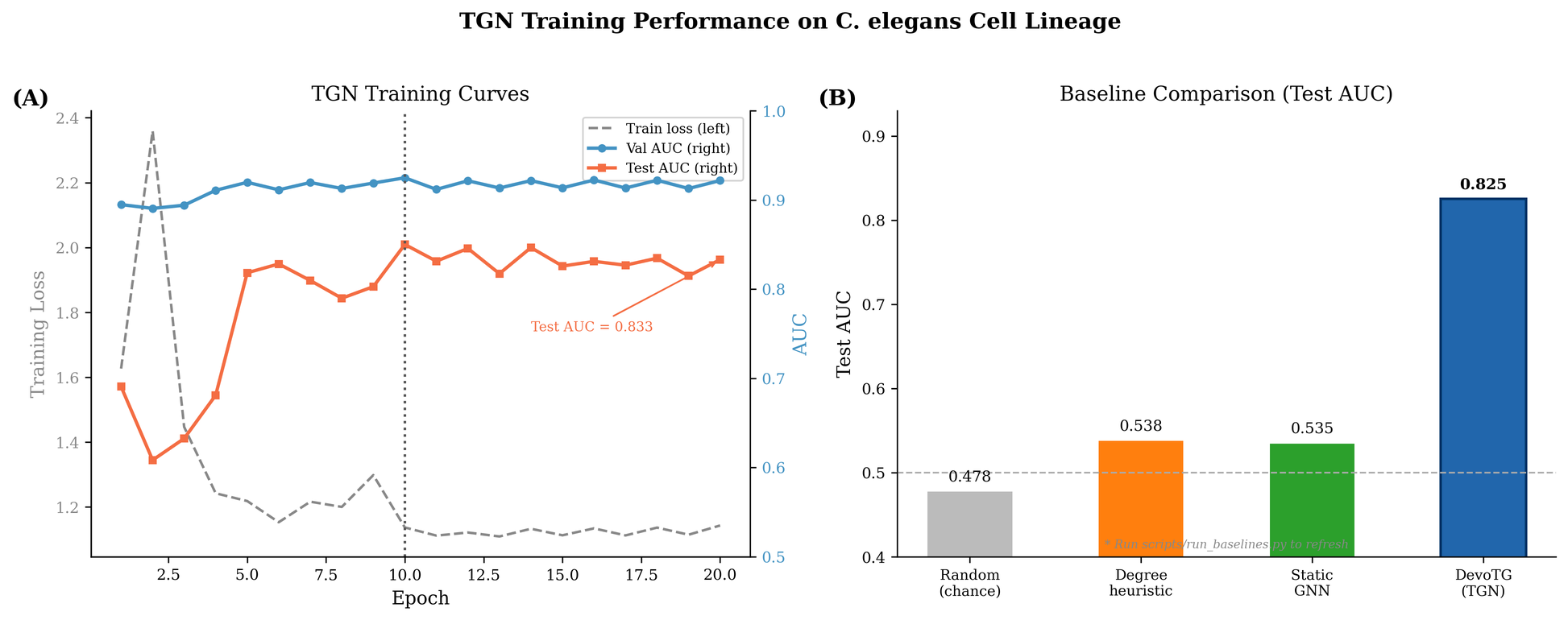}\hfill
  \includegraphics[width=0.48\linewidth]{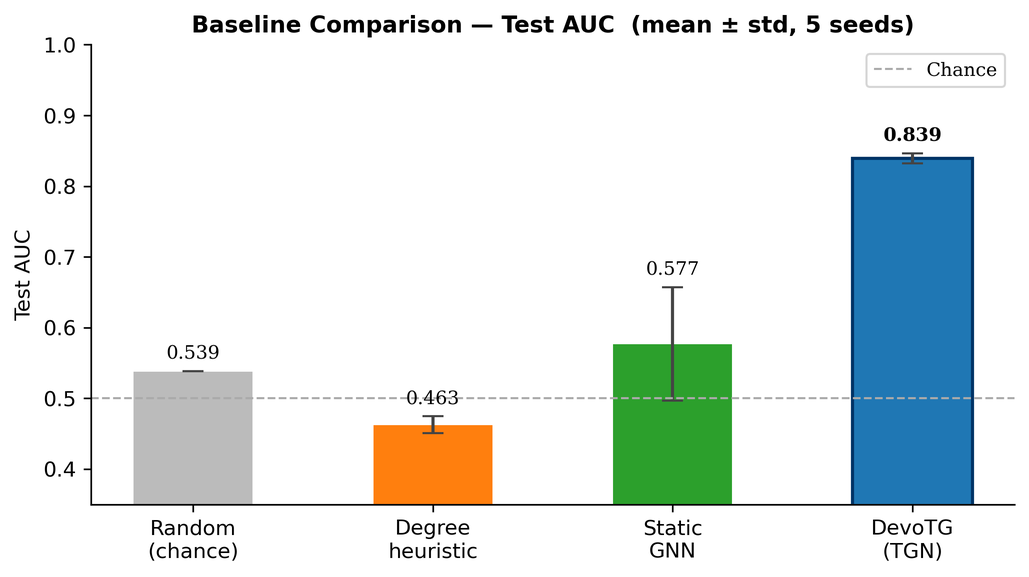}
  \caption{\textbf{TGN training performance and multi-seed baseline comparison.}
    Left: Training curves over 20 epochs (single representative run, seed 0).
    The best checkpoint is selected by validation AUC; test AUC = 0.839.
    Right: Test AUC mean $\pm$ std across 5 seeds for all four models.
    Error bars show seed-to-seed variability.
    The 26-point gap between Static GNN and TGN, with low TGN variance
    ($\pm$0.007), confirms that temporal memory is the decisive factor.}
  \label{fig:performance}
\end{figure}

The best checkpoint for each seed is selected by validation AUC (mean
$0.937 \pm 0.001$), following standard model-selection practice; only test
metrics are reported in Table~\ref{tab:baselines}.
Figure~\ref{fig:performance} shows training dynamics (A) and the multi-seed
baseline comparison with error bars (B).

\section{Discussion}
\label{sec:discussion}
\paragraph{Temporal memory is indispensable for developmental prediction.}
The \textbf{26-point test AUC gap} between DevoTG's TGN ($0.839 \pm 0.007$)
and the Static GNN baseline ($0.577 \pm 0.080$), measured over 5 random seeds,
is the most striking quantitative result of this work.
Both models share the same TransformerConv graph attention architecture; the
sole difference is the TGN memory module, a GRU that accumulates each cell's
interaction history as a 100-dimensional state vector.
The TGN's low variance ($\pm$0.007 across seeds) contrasts with the Static
GNN's instability ($\pm$0.080) further confirms that temporal structure,
not lucky initialization, drives the TGN's advantage.
This result establishes that the \textit{order} of cell division events, not
just their occurrence, encodes information critical for predicting future
divisions.
In biological terms, a cell's developmental competence depends on its lineage
history: the same cell type, born at different stages or from different
lineages, will have different proliferative fates.
The TGN memory module provides an inductive bias aligned with this biological
reality in a way that static graph methods cannot.

\paragraph{Temporal and individual variability are complementary.}
DevoTG's temporal stability classification (stable: 15.1\%, developmental:
28.1\%, variable: 56.8\%) and Witvliet's individual-variability classification
(stable+dynamic $\approx$57\%, variable $\approx$43\%) measure different
but related properties of the connectome.
Witvliet's variable connections are unique to each worm; DevoTG's variable
connections appear in only one of eight developmental timepoints.
The two axes are not independent: connections that are weak and transient in
the temporal dimension are also more likely to be absent in some individuals.
Future work integrating multiple animals at each timepoint would allow direct
comparison, and DevoTG's framework is designed to accommodate such an extension
by treating each animal as a parallel DTDG instance.

\paragraph{AVA/AVB/AVE as temporal anchors.}
The finding that AVA, AVB, and AVE show strong inter-neuron connectivity
from birth, even as their neighborhoods expand substantially by adulthood,
supports Witvliet's conclusion that ``the central decision-making circuitry
is maintained'' while sensory and motor pathways remodel.
DevoTG adds temporal resolution to this claim: the command interneuron core
is not just preserved but progressively reinforced, with new connections from
downstream motor neurons and upstream sensory neurons adding integration
capacity over the course of larval development.
This reinforcement pattern, visible in Figure~\ref{fig:circuit}, is consistent
with the behavioral observation that \textit{C.~elegans} locomotion becomes
more coordinated and responsive with age.

\paragraph{Limitations and future directions.}
Several limitations warrant acknowledgment.
First, the TGN is trained on cell lineage data, which captures cell division
events but not the continuous biological processes (axon outgrowth, synapse
formation, and activity-dependent plasticity) that drive connectivity.
Training the TGN directly on the connectome DTDG (predicting synapse formation
between timepoints) is a natural next step.
Second, the stability classification uses a fixed threshold (6/8 timepoints
for ``stable''), and sensitivity to this threshold should be evaluated.
Third, gap junctions are treated equivalently to chemical synapses in the
DTDG; their different biophysical properties and developmental regulation
may warrant separate analysis.
Fourth, the current analysis is restricted to \textit{C.~elegans};
comparative developmental connectomics across species would test the
generality of the observed maturation principles.

Despite these limitations, DevoTG demonstrates that temporal graph methods
can extract biologically meaningful structure from developmental connectome
data, and provides a foundation for more comprehensive analysis as
longitudinal connectome datasets grow in availability.

\section{Conclusion}
\label{sec:conclusion}
We introduced DevoTG, a temporal graph framework for computational analysis of
\textit{C.~elegans} neural development.
Through two complementary representations (a CTDG of cell division events and
a DTDG of the developing synaptic connectome), DevoTG enables both predictive
modeling and descriptive analysis of developmental connectivity dynamics.

On the cell division prediction task, our TGN achieves a mean test AUC of
$0.839 \pm 0.007$ (5 seeds), 26 points above a static GNN baseline with the
same architecture ($0.577 \pm 0.080$), demonstrating that temporal memory
provides a critical inductive bias that static graph methods cannot replicate.
Applied to the Witvliet connectome, DevoTG identifies three connection
stability classes whose distributions inform and complement the
individual-variability taxonomy of Witvliet et al., and reveals the
progressive reinforcement of command interneuron hub status across larval
development.

DevoTG is available as open-source software with interactive visualizations,
modular data loaders, and runnable analysis scripts.
We hope it serves as a foundation for the emerging field of temporal
connectomics, enabling systematic, data-driven study of how nervous systems
wire themselves from birth to maturity.

\section*{Acknowledgments}
This work was supported by Google Summer of Code 2025, in collaboration with the International Neuroinformatics Coordinating Facility (INCF; \url{https://www.incf.org}). We thank Bradly Alicea and Mehul Arora for their mentorship and guidance throughout the project.
\bibliographystyle{plainnat}
\bibliography{references}

@article{witvliet2021connectomes,
  title={Connectomes across development reveal principles of brain maturation},
  author={Witvliet, Daniel and Mulcahy, Ben and Mitchell, James K and Meirovitch, Yaron and Berger, Daniel R and Wu, Yuelong and Liu, Yufang and Koh, Wan Xian and Parvathala, Rajeev and Holmyard, Douglas and others},
  journal={Nature},
  volume={596},
  number={7871},
  pages={257--261},
  year={2021},
  publisher={Nature Publishing Group UK London}
}

@article{white1986structure,
  title={The structure of the nervous system of the nematode Caenorhabditis elegans: the mind of a worm},
  author={White, John G and Southgate, Eileen and Thomson, J Nichol and Brenner, Sydney},
  journal={Phil. Trans. R. Soc. Lond},
  volume={314},
  number={1},
  pages={340},
  year={1986}
}

@inproceedings{rossi2020temporal,
    title={Temporal Graph Networks for Deep Learning on Dynamic Graphs},
    author={Emanuele Rossi and Ben Chamberlain and Fabrizio Frasca and Davide Eynard and Federico 
    Monti and Michael Bronstein},
    booktitle={ICML 2020 Workshop on Graph Representation Learning},
    year={2020}
}

@article{alicea2020raising,
  title={Raising the connectome: the emergence of neuronal activity and behavior in \textit{Caenorhabditis elegans}},
  author={Alicea, Bradly},
  journal={Frontiers in Cellular Neuroscience},
  volume={14},
  pages={290},
  year={2020},
  publisher={Frontiers Media SA}
}

@article{frankel2025devsynapse,
  title={Principles of synaptogenesis: Insights from \textit{Caenorhabditis elegans}.},
  author={Frankel, Elisa B. and Kurshan, Peri T.},
  journal={Current Opinion in Neurobiology},
  volume={93},
  pages={103056},
  year={2025},
  publisher={Elsevier}
}

@article{alicea2018spatialnetworks,
  title={Cell Differentiation Processes as Spatial Networks: identifying four-dimensional structure in embryogenesis},
  author={Alicea, Bradly and Gordon, Richard},
  journal={Biosystems},
  volume={173},
  pages={235-246},
  year={2018},
  publisher={Elsevier}
}

@inproceedings{beck2017visual,
  title={A taxonomy and survey of dynamic graph visualization},
  author={Beck, Fabian and Burch, Michael and Diehl, Stephan and Weiskopf, Daniel},
  booktitle={Computer Graphics Forum},
  volume={36},
  pages={133--159},
  year={2017},
  organization={Wiley Online Library}
}

@article{longa2023graph,
  title={Graph neural networks for temporal graphs: State of the art, open challenges, and opportunities},
  author={Longa, Antonio and Lachi, Veronica and Santin, Gabriele and Bianchini, Monica and Lepri, Bruno and Lio, Pietro and Scarselli, Franco and Passerini, Andrea},
  journal={arXiv preprint arXiv:2302.01018},
  year={2023}
}

@article{bassett2017network,
  title={Network neuroscience},
  author={Bassett, Danielle S and Sporns, Olaf},
  journal={Nature Neuroscience},
  volume={20},
  number={3},
  pages={353--364},
  year={2017},
  publisher={Nature Publishing Group US New York}
}

@article{trivedi2019dyrep,
  title={Dyrep: Learning representations over dynamic graphs},
  author={Trivedi, Rakshit and Farajtabar, Mehrdad and Biswal, Prasenjeet and Zha, Hongyuan},
  journal={International Conference on Learning Representations},
  year={2019}
}

@article{hajiramezanali2019variational,
  title={Variational graph recurrent neural networks},
  author={Hajiramezanali, Ehsan and Hasanzadeh, Arman and Narayanan, Krishna and Duffield, Nick and Zhou, Mingyuan and Qian, Xiaoning},
  journal={Advances in neural information processing systems},
  volume={32},
  year={2019}
}

@article{kumar2020predicting,
  title={Predicting dynamic embedding trajectory in temporal interaction networks},
  author={Kumar, Srijan and Zhang, Xikun and Leskovec, Jure},
  journal={Proceedings of the 25th ACM SIGKDD international conference on knowledge discovery \& data mining},
  pages={1269--1278},
  year={2020}
}

@article{hosseinzadeh2022temporal,
  title={Temporal networks in biology and medicine: a survey on models and algorithms},
  author={Hosseinzadeh, Mohammad Mehdi and Eftekhari, Mostafa and Sadoughi, Farahnaz},
  journal={Artificial Intelligence in Medicine},
  volume={119},
  pages={102157},
  year={2022},
  publisher={Elsevier}
}

@article{watts1998collective,
  title={Collective dynamics of 'small-world' networks},
  author={Watts, Duncan J and Strogatz, Steven H},
  journal={nature},
  volume={393},
  number={6684},
  pages={440--442},
  year={1998},
  publisher={Nature Publishing Group}
}

@article{towlson2013caenorhabditis,
  title={The rich club of the \textit{C. elegans} neuronal connectome},
  author={Towlson, Emma K and V{\'e}rtes, Petra E and Ahnert, Sebastian E and Schafer, William R and Bullmore, Edward T},
  journal={Journal of Neuroscience},
  volume={33},
  number={15},
  pages={6380--6387},
  year={2013},
  publisher={Soc Neuroscience}
}

@article{holme2012temporal,
  title={Temporal networks},
  author={Holme, Petter and Saram{\"a}ki, Jari},
  journal={Physics Reports},
  volume={519},
  number={3},
  pages={97--125},
  year={2012},
  publisher={Elsevier}
}

@article{rosvall2010mapping,
  title={Mapping change in large networks},
  author={Rosvall, Martin and Bergstrom, Carl T},
  journal={Proceedings of the national academy of sciences},
  volume={107},
  number={13},
  pages={5825--5830},
  year={2010},
  publisher={National Acad Sciences}
}

@article{varshney2011structural,
  title={Structural properties of the Caenorhabditis elegans neuronal network},
  author={Varshney, Lav R and Chen, Beth L and Paniagua, Eric and Hall, David H and Chklovskii, Dmitri B},
  journal={PLOS Computational Biology},
  volume={7},
  number={2},
  pages={e1001066},
  year={2011},
  publisher={Public Library of Science}
}

@article{cook2019whole,
  title={Whole-animal connectomes of both Caenorhabditis elegans sexes},
  author={Cook, Steven J and Jarrell, Travis A and Bhatt, Christopher A and others},
  journal={Nature},
  volume={571},
  number={7763},
  pages={63--71},
  year={2019},
  publisher={Nature Publishing Group}
}

@article{pareja2020evolvegcn,
  title={{EvolveGCN}: Evolving graph convolutional networks for dynamic graphs},
  author={Pareja, Aldo and Domeniconi, Giovanni and Chen, Jie and Ma, Tengfei and Suzumura, Toyotaro and Kanezashi, Hiroki and Kaler, Tim and Schardl, Tao and Leiserson, Charles},
  journal={Proceedings of the AAAI Conference on Artificial Intelligence},
  volume={34},
  number={04},
  pages={5363--5370},
  year={2020}
}

\appendix
\section*{Appendix}

\subsection*{A. TGN Hyperparameters and Training Details}

\begin{table}[h]
\centering
\caption{TGN hyperparameters and training configuration.}
\small
\begin{tabular}{ll}
\toprule
\textbf{Model architecture} & \\
\midrule
Memory dimension $d_m$ & 100 \\
Time encoding dimension $d_t$ & 100 \\
Embedding dimension $d_e$ & 100 \\
TransformerConv heads & 2 \\
TransformerConv dropout & 0.1 \\
Message module & IdentityMessage \\
Aggregator module & LastAggregator \\
Temporal neighbors per node & 10 (LastNeighborLoader) \\
Total trainable parameters & 132{,}501 \\
\midrule
\textbf{Training} & \\
\midrule
Optimizer & Adam \\
Learning rate & 0.001 \\
Loss function & BCEWithLogitsLoss \\
Batch size & 200 \\
Negative sampling ratio & 1:1 (per batch) \\
Epochs & 20 \\
Train / Val / Test split & 70\% / 15\% / 15\% (temporal) \\
Checkpoint selection & Best validation AUC \\
Evaluation seeds & 5 (seeds 0--4) \\
\midrule
\textbf{Hardware} & \\
\midrule
GPU & NVIDIA GeForce GTX 1650, 4 GB GDDR5 \\
Training time & $\sim$3 s per seed (20 epochs) \\
\bottomrule
\end{tabular}
\end{table}

\subsection*{B. Static GNN Baseline Architecture}

The static GNN baseline uses two \texttt{GATConv} layers
(PyTorch Geometric implementation):
\begin{itemize}
  \item Layer 1: input 172 $\rightarrow$ 50 per head, 2 heads (concatenated),
        output 100; ELU activation; dropout 0.1.
  \item Layer 2: input 100 $\rightarrow$ 100, 1 head (no concatenation);
        ELU activation; dropout 0.1.
\end{itemize}
The static adjacency matrix is built from all training events (undirected).
The same \texttt{LinkPredictor} head as the TGN is used for scoring.
Training uses the same optimizer, loss, batch size, and epochs as the TGN.
The mean test AUC ($0.577 \pm 0.080$ over 5 seeds) confirms that this is a
competitive, well-tuned baseline whose underperformance demonstrates the
importance of temporal memory.

\subsection*{C. Connectome Centrality Analysis}

The top five nodes by degree centrality in the adult connectome (averaged
across the two adult timepoints, datasets 7 and 8) are:

\begin{center}
\small
\begin{tabular}{lll}
\toprule
Neuron & Type & Degree Centrality \\
\midrule
RIML & inter & 0.252 \\
AIBL & inter & 0.252 \\
AIBR & inter & 0.243 \\
AVEL & inter & 0.239 \\
RMDL & inter & 0.234 \\
\bottomrule
\end{tabular}
\end{center}

All top-centrality neurons are interneurons, consistent with their role as
hub integrators.
The command interneurons AVAL/AVAR (not in the top 5 by simple degree
centrality) have significantly higher \emph{betweenness} centrality, reflecting
their roles as conduits between sensory and motor layers.

\subsection*{D. Code and Data Availability}

Code is publicly available at \url{https://github.com/DevoLearn/DevoGraph/tree/main/DevoTG} (DevoLearn organization repository). An updated, actively maintained version is available at \url{https://github.com/Jayadratha/DevoTG_GSoC} (branch: research).
Connectome data: Witvliet et al.\ 2021 (Nature), accessible at
\url{http://www.nemanode.org} and via the ConnectomeToolbox repository.
Cell lineage data: WormAtlas canonical lineage, processed CSV included in the
repository.
All analysis scripts and figure generation code are provided with environment
specification (\texttt{environment.yml}, Python 3.12, PyTorch 2.5.1,
PyG 2.6.1).

\end{document}